\documentclass[10.5pt,compsoc]{CCSIG}
\usepackage{graphicx}
\usepackage{footmisc}
\usepackage{subfigure}
\usepackage{url}
\usepackage{multirow}
\usepackage[noadjust]{cite}
\usepackage{amsmath,amsthm}
\usepackage{amssymb,amsfonts}
\usepackage{booktabs}
\usepackage{color}
\usepackage{ccaption}
\usepackage{booktabs}
\usepackage{float}
\usepackage{fancyhdr}
\usepackage{caption}
\usepackage{xcolor,stfloats}
\usepackage{comment}
\usepackage{cuted}
\usepackage{captionhack}
\usepackage{epstopdf}
\usepackage{tabularx}
\usepackage{array}

\headevenname{\mbox{\quad} \hfill  \mbox{ChinaMFS 2025}}%
\headoddname{
ChinaMFS 2025\hfill}%

\newtheoremstyle{mystyle}{0pt}{0pt}{\normalfont}{1em}{\bf}{}{1em}{}
\theoremstyle{mystyle}
\renewcommand\figurename{figure~}

\makeatletter
\renewcommand{\@biblabel}[1]{[#1]\hfill}
\makeatother
\begin{document}

\hyphenpenalty=50000
\makeatletter
\newcommand\mysmall{\@setfontsize\mysmall{7}{9.5}}
\newenvironment{tablehere}
  {\def\@captype{table}}

\let\temp\footnote
\renewcommand \footnote[1]{\temp{\zihao{-5}#1}}

\thispagestyle{plain}%
\thispagestyle{empty}%
\pagestyle{CCSIGheadings}

\begin{table*}[!t]
\vspace {-13mm}
\begin{tabular}{p{168mm}}
\hfill The 6th CSIG Chinese Conference on Media Forensics and Security \quad\quad\quad\quad\quad\quad\quad\quad\quad\quad\quad\quad\quad\\
\hline\\[-4.5mm]
\hline\end{tabular}

\vskip 7mm

\begin{center}
Hybrid Retrieval-Augmented Generation Agent for Trustworthy Legal Question Answering in Judicial Forensics \\
\vspace {5mm}
Yueqing Xi$^{1)}$  Yifan Bai$^{2)}$ Huasen Luo$^{1)}$ Weiliang Wen$^{1)}$ Hui Liu $^{2)}$Haoliang Li$^{1)}$ $^{2)}$ *\\
\vspace {2mm}
$^{1)}$(Department of Electronic Engineering, City University of Hong Kong(DongGuan), DongGuan)\\
$^{2)}$(Department of Electronic Engineering, City University of Hong Kong, HongKong)\\

\end{center}

\begin{tabular}{p{160mm}}
\zihao{5}{
\setlength{\baselineskip}{18pt}\selectfont{
{\bf Abstract}
\par}}\\

\setlength{\baselineskip}{18pt}\selectfont{
\zihao{5}{\noindent As artificial intelligence permeates judicial forensics, ensuring the veracity and traceability of legal question answering (QA) has become critical. Conventional large language models (LLMs) are prone to hallucination, risking misleading guidance in legal consultation, while static knowledge bases struggle to keep pace with frequently updated statutes and case law. We present a hybrid legal QA agent tailored for judicial settings that integrates retrieval-augmented generation (RAG) with multi-model ensembling to deliver reliable, auditable, and continuously updatable counsel. The system prioritizes retrieval over generation: when a trusted legal repository yields relevant evidence, answers are produced via RAG; otherwise, multiple LLMs generate candidates that are scored by a specialized selector, with the top-ranked answer returned. High-quality outputs then undergo human review before being written back to the repository, enabling dynamic knowledge evolution and provenance tracking. Experiments on the Law\_QA dataset show that our hybrid approach significantly outperforms both a single-model baseline and a vanilla RAG pipeline on F1, ROUGE-L, and an LLM-as-a-Judge metric. Ablations confirm the complementary contributions of retrieval prioritization, model ensembling, and the human-in-the-loop update mechanism. The proposed system demonstrably reduces hallucination while improving answer quality and legal compliance, advancing the practical landing of media forensics technologies in judicial scenarios.

\vspace {5mm}
{\bf Keywords}: Judicial Forensics, Legal Question Answering, Retrieval-Augmented Generation(RAG), Multi-Model Ensembling, Trustworthy Generation, Knowledge Rewriting and Updating}\par}
\end{tabular}

\setlength{\tabcolsep}{2pt}
\begin{tabular}{p{0.05cm}p{16.15cm}}
\multicolumn{2}{l}{\rule[4mm]{40mm}{0.1mm}}\\[-3mm]
&
\end{tabular}\end{table*}

\clearpage\clearpage
\begin{strip}
\vspace {-13mm}
\end{strip}
    \linespread{1.15}
\vskip 1mm

\section{Introduction}
With the rapid advancement of artificial intelligence, generative large language models (LLMs) are being increasingly applied in the legal domain—especially in judicial forensics—opening new possibilities for legal consultation, evidence analysis, and case retrieval\cite{re2,re7,re8}. However, judicial settings impose stringent requirements on authenticity and traceability. Conventional LLMs are prone to hallucinations—i.e., producing unsupported or inaccurate content—which poses significant risks in legal consultation\cite{re3}. For instance, mistakes or fabrications in statutory citations or case analyses can lead to serious judicial misjudgments, directly undermining the fairness of adjudication. Moreover, legal knowledge evolves rapidly: statutes and case law are frequently updated, and static knowledge bases struggle to keep pace, limiting the practical deployment of AI systems in real judicial scenarios\cite{re4}.

To address these challenges, retrieval-augmented generation (RAG) has been widely explored to improve answer quality by combining LLMs with external and
citable knowledge sources\cite{re1,re5,re21}. By retrieving relevant information from structured repositories, RAG grounds generation in verifiable evidence and markedly reduces hallucinations. Yet a single RAG pipeline may still fail on complex or novel legal questions when the knowledge base lacks coverage. In addition, many existing legal-RAG systems are static, lacking mechanisms for dynamic updates that reflect the fast-changing needs of the judicial domain \cite{re4,re17,re18}. On the other hand, multi-model ensembling aggregates responses from multiple LLMs and selects among them, improving robustness and reliability; nevertheless, such methods can be computationally expensive and require further optimization in judicial applications to ensure terminological precision and explainability\cite{re6,re16}.

Against this backdrop, we propose a hybrid legal QA agent tailored to judicial scenarios, aiming for high reliability, traceability, and continual knowledge refresh. Our system combines RAG with multi-model ensembling: when a user query matches high-similarity entries in the knowledge base, we answer via RAG to ensure grounding in validated legal sources; when retrieval misses, we generate candidate answers from multiple pretrained models and apply a specialized selector to score and elect the best response. Furthermore, high-quality answers undergo human review and are then written back into the knowledge base, enabling dynamic evolution aligned with changes in the law. Experiments demonstrate that our approach outperforms both traditional baselines and single-RAG pipelines on legal QA datasets; ablation studies further verify the contribution of each module, showcasing its potential for judicial forensics applications\cite{re15,re23}.

Our main contributions are threefold: (i) we design and implement a judicial-specific hybrid agent that marries RAG with multi-model ensembling to improve the fidelity and reliability of legal QA; (ii) we introduce a dynamic knowledge-base update mechanism, combining human review with automatic scoring to ensure long-term validity and compliance; and (iii) we empirically validate the superiority of our system on legal QA tasks, offering new perspectives for industry deployment in digital media forensics. Aligned with the conference theme, this work explores authenticity verification and safety protection for AI in judicial forensics—particularly the standardization and compliance of legal consultation—providing both theoretical and practical references for intelligent judicial systems.

The remainder of this paper is organized as follows: Section 2 reviews related work, covering digital media forensics and the application of RAG and ensembling in judicial contexts; Section 3 details the proposed system architecture and methods; Section 4 presents experimental setup and results; Section 5 discusses advantages, limitations, and application prospects; and Section 6 concludes and outlines future directions.

\section{Related Work}
Digital media forensics and security, as a research hot spot at the intersection of artificial intelligence and the legal domain, have shown significant value in scenarios such as legal consultation, evidence analysis, and case retrieval in recent years. This paper focuses on the judicial landing of forensics technologies and proposes a reliable legal QA agent system by integrating retrieval-augmented generation (RAG), multi-model ensembling, and a dynamic knowledge-base updating mechanism. We review related progress from five perspectives—digital media forensics, RAG in legal QA, ensemble methods, AI deployment in judicial contexts, and dynamic knowledge-base updating—then highlight their relation to and gaps from our work.

\subsection{Digital Media Forensics}
Digital media forensics aims to verify the authenticity and integrity of digital content through technical means, widely applied in judicial scenarios for evidence analysis and forgery detection. For example, with the rapid development of deepfake techniques, researchers have proposed neural network–based detection algorithms to identify manipulated video or textual evidence\cite{re10,re11,re22}. In the legal domain, AI methods have been employed for automated evidence collection and analysis, such as extracting key information from large-scale legal documents using NLP. However, existing forensics systems often emphasize multi-modal data (images, video) rather than textual legal QA. Moreover, they remain vulnerable to LLM hallucinations\cite{re3}. Our hybrid agent system fills this gap by combining RAG and ensemble strategies to focus on textual authenticity verification in legal QA.

\subsection{RAG in Legal Question Answering}
RAG improves LLM factuality by grounding answers in retrieved external knowledge. In legal QA, this has demonstrated clear advantages\cite{re1,re21}. \cite{re12} introduced the CBR-RAG framework, which integrates case-based reasoning with RAG to enhance semantic accuracy by retrieving similar past cases; experiments showed ~20\% reduction in hallucinations. \cite{re13} developed LexRAG, a benchmark dataset for multi-turn legal consultations, finding that retrieval coverage remains limited for novel legal issues. Barron et al. further incorporated knowledge graphs and non-negative matrix factorization (NMF) to optimize retrieval efficiency for semi-structured legal data\cite{re14}, but their static design cannot adapt to rapidly evolving statutes. In contrast, our work augments RAG with a dynamic updating mechanism and hybrid fallback strategies to improve adaptability in judicial contexts.

\subsection{Multi-Model Ensembling}
Ensembling aggregates the outputs of multiple LLMs and applies ranking or voting, thereby enhancing robustness and reliability\cite{re6,re16}. \cite{re12} proposed a hybrid fuzzy logic-random forest model to predict psychiatric case outcomes, with high accuracy and interpretability critical for transparency. Similarly, Khan et al. designed a legal AI framework combining expert systems and RAG, demonstrating reduced hallucinations in complex legal tasks\cite{re13}. Yet, most existing ensemble systems are resource-intensive and insufficiently optimized for terminological precision in law. Our approach introduces a specialized selector model that scores and elects among multiple outputs, combined with RAG prioritization, balancing efficiency and legal accuracy.

\subsection{AI Deployment in Judicial Scenarios}
Practical deployment of AI in judicial contexts requires reliability, interpretability, and compliance. \cite{re7,re9} studied legal AI systems integrating RAG and knowledge graphs, emphasizing scalability and cost-effectiveness but overlooking dynamic knowledge updating. Current judicial AI mainly focuses on case retrieval or judgment prediction, with less attention to dynamic QA updating and compliance in forensics. Our system strengthens compliance and long-term effectiveness through dynamic updating plus human review, aligning with industry needs for standardization and regulatory adherence.

\subsection{Dynamic Knowledge-Base Updating}
Legal knowledge changes frequently—new statutes and precedents must be rapidly incorporated. \cite{re14} explored clustering and updating legal knowledge via NMF and web scraping, which improved scalability but introduced risks due to data quality and compliance. \cite{re13} proposed reinforcement learning with human feedback (RLHF) to enhance legal knowledge-base updating for smart courts, though the human-review cost was not fully addressed. Our update mechanism combines automatic scoring with human review, balancing efficiency and compliance. By writing high-quality reviewed answers back into the repository, our system ensures that the knowledge base evolves reliably with the law, particularly important for long-running forensic applications\cite{re23}.

\subsection{Gap Analysis}
While prior research in RAG, ensembling, and knowledge updating offers valuable insights, limitations remain. (i) Single RAG systems struggle with novel queries due to knowledge-base coverage gaps; (ii) Ensemble methods improve robustness but are resource-heavy and insufficiently tuned for legal precision; (iii) Current knowledge bases are mostly static or semi-automatically updated, failing to meet the dynamic demands of judicial domains. Our hybrid agent addresses these shortcomings by combining RAG, ensembling, and dynamic updating into a unified pipeline, achieving high reliability, traceability, and adaptability, and advancing authenticity verification and security in judicial forensics.

\section{The Proposed Method}
This paper proposes a hybrid legal question-answering (QA) agent system tailored for judicial scenarios. The system integrates retrieval-augmented generation (RAG)\cite{re1} with multi-model ensembling to deliver reliable, traceable, and dynamically updatable legal consultation services. Specifically, the system prioritizes generating answers based on a trusted knowledge base; when retrieval fails, it resorts to multi-model generation and selector-based ranking to ensure accuracy and compliance. High-quality answers are further reviewed by humans and dynamically updated into the knowledge base, enabling adaptation to the fast evolution of legal knowledge. This section details the overall architecture, knowledge base construction, RAG generation, multi-model mechanism, and dynamic updating workflow.

\subsection{System Architecture Overview}
\begin{figure}[htbp]
\centering
\includegraphics[width=0.4\textwidth]{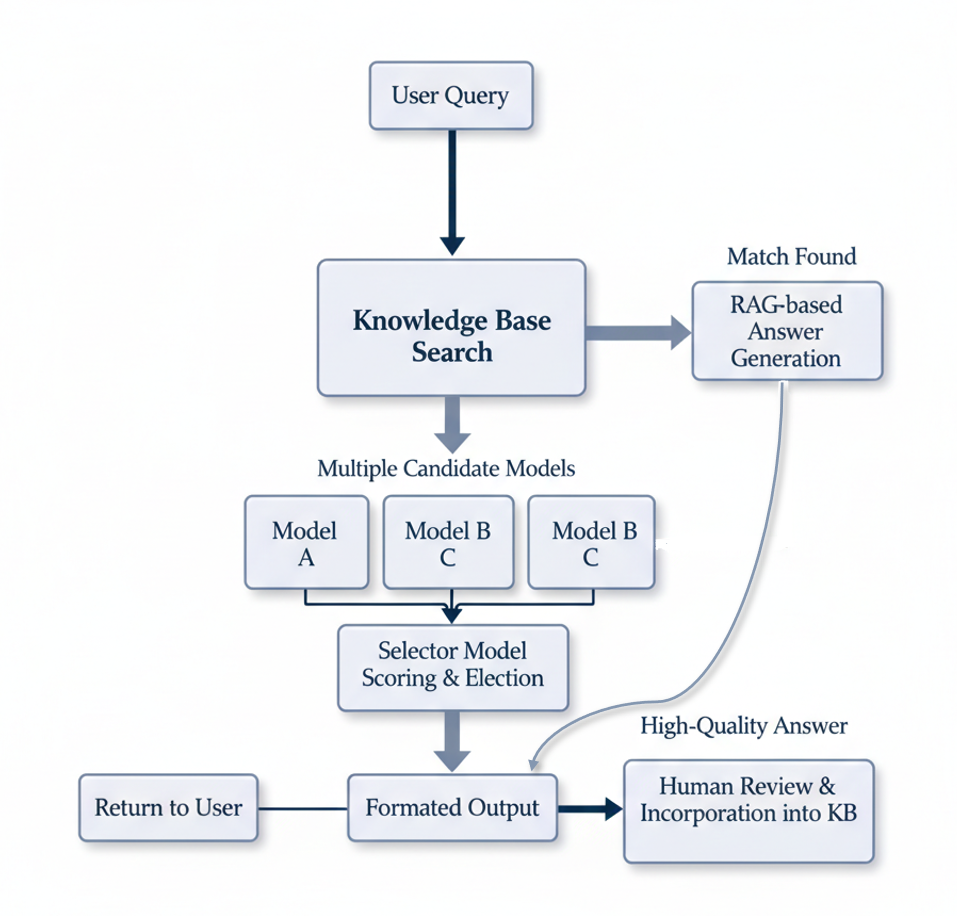}
\caption{\;Framework of the proposed legal query answering system.}
\label{fig:1}
\end{figure}

At the core lies a hybrid agent responsible for processing legal queries and generating reliable answers. Fig.\ref{fig:1} illustrates the workflow: upon receiving a user query, the system first searches the knowledge base for relevant entries. If a match is found, the system applies RAG to generate an answer. If no match is found, multiple models are invoked to produce candidate answers, which are then scored by a specialized selector model to elect the best one. The final output is formatted and returned to the user, while high-quality answers undergo human review and are incorporated into the knowledge base. This pipeline ensures authenticity, traceability, and long-term adaptability, particularly addressing the high-reliability demands of judicial forensics. 

\subsection{Reliable Knowledge Base Construction}
The knowledge base serves as the foundation for authenticity verification. It is designed in a structured manner to store legal QA data. We use the m3e-base text embedding model to encode both questions and candidate answers\cite{re20}, generating text embeddings mapped via FAISS for efficient vector retrieval\cite{re19}. Each entry is stored in the format:(id,question,answer,cause), where question and answer represent legal queries and their standard responses, while cause provides legal basis (e.g., statutes or precedents). To reduce runtime overhead, all QA entries are preloaded into memory as dictionary rows, with cached files recording embedding–index mappings. Batch retrieval supports multi-query processing, and cosine similarity is used to compute query–entry relevance. The system returns top-k results after deduplication, ensuring both diversity and accuracy of retrieval.

\subsection{RAG Generation Process}
When the similarity between a user query and knowledge base entries exceeds the threshold (set to 0.6 in this paper), the system adopts the RAG pipeline. The RAG prompt consists of system instructions and user templates: the instructions require the model to generate a natural 2–4 sentence response based on retrieved entries, explicitly citing statutory provisions or regulations to ensure compliance. A single pretrained model then generates the answer. The output undergoes post-processing—removing code blocks, normalizing whitespace, and constraining length ($\leq 280$ characters)—to meet the conciseness requirements of judicial consultation.

\subsection{Multi-Model Hybrid Mechanism}
If no high-similarity entry is retrieved, the system switches to the multi-model fallback mechanism for novel or complex legal queries:

1.Parallel Generation: We employ three pretrained models, ChatGPT-4o, Qwen3-235B-A22B, and DeepSeek-v3.1 to simultaneously produce candidate answers. The generation is guided by a baseline template, ensuring consistency with the natural and formal style required in judicial QA.

2.Selector-Based Scoring: A specialized selector model (e.g., Google Gemini-2.5-flash-lite) evaluates candidates along five dimensions: correctness, legality, completeness, clarity, and fidelity. The candidate with the highest score is selected as the system output.

This mechanism reduces the risk of hallucinations through multi-model collaboration, while selector-based ranking enhances reliability, particularly when addressing queries beyond the coverage of the knowledge base.

\section{Experiment}
This section introduces the experimental setup and results for the proposed hybrid legal QA agent in judicial scenarios. We evaluate system performance on the LawQA dataset to validate its effectiveness in legal forensics. By comparing the baseline model, RAG-enhanced model, and hybrid model (RAG + multi-model ensembling), together with ablation studies and analyses of knowledge-base indexing strategies, we demonstrate the system’s advantages in accuracy, traceability, and adaptability. Results show that the hybrid model outperforms both the baseline and standalone RAG approaches in terms of F1 score, ROUGE score, and LLM-as-a-Judge rating, while ablation experiments further verify the contributions of each module.

\subsection{Experimental Setup}
\subsubsection{Experimental Setup}
The experiments are conducted on the Law\_QA dataset, which contains legal QA pairs from Mainland China across multiple domains, including civil law, labor law, and marriage law, making it suitable for evaluating forensic QA systems in judicial contexts. The dataset consists of 16,182 QA pairs, split into training and validation sets at an 8:2 ratio, with three-fold cross-validation. The training set is used to construct the reliable knowledge base (including FAISS indexing and vector embeddings), while the validation set is used to evaluate system performance. Each QA pair includes the question, standard answer, and legal basis (cause), ensuring both traceability and compliance.

\subsubsection{Model Configuration}
We adopt the following configurations:
Baseline model: A single pretrained model (ChatGPT-4o in this study), generating answers using a baseline prompt template.
RAG model: Combines knowledge-base retrieval (using m3e-base embeddings with FAISS indexing) and RAG prompting. If retrieval succeeds, ChatGPT-4o generates an answer grounded in the retrieved QA pairs.
Hybrid model: Extends RAG with a fallback multi-model ensembling strategy. For queries with no retrieval match, three pretrained models (ChatGPT-4o, Qwen3-235B-A22B, DeepSeek-v3.1) generate candidate answers, which are then ranked by a selector model (Google Gemini-2.5-flash-lite).
Selector and Judge: Google Gemini-2.5-flash-lite serves as both the selector and evaluation model. It scores candidate answers based on dimensions including correctness, legality, completeness, clarity, and faithfulness, with the highest scoring answer selected for the output.

\subsubsection{Evaluation Metrics}
We adopt three evaluation metrics, F1 score, ROUGE-L, and LLM-as-a-Judge, in line with previous work.
F1 Score is a widely used comprehensive evaluation metric in information retrieval and natural language processing. It evaluates system performance by calculating the harmonic mean of Precision and Recall. In this study, we adopt character-level F1 score ($F1_{char}$), whose formula is as follows.\\
Precision: measures the proportion of correct characters in the predicted answer
\begin{equation}
Precision =\frac{TP}{TP+FP}
\end{equation}
Recall: measures the proportion of characters in the standard answer that are correctly predicted:
\begin{equation}
Recall=\frac{TP}{TP+FN}
\end{equation}
F1 Score: the harmonic mean of Precision and Recall:
\begin{equation}
F1=2\times\frac{Precision\times Recall}{Precision+Recall}
\end{equation}
Here, TP (True Positive) denotes the number of correctly predicted characters, FP (False Positive) denotes the number of incorrectly predicted characters, and FN (False Negative) denotes the number of correct characters that were not predicted. The character-level F1 score can capture semantic similarity more finely, making it particularly suitable for evaluating Chinese legal text.

ROUGE (Recall-Oriented Understudy for Gisting Evaluation) is a commonly used metric for text summarization and quality evaluation. In this study, we adopt the ROUGE-L metric, which is based on the Longest Common Subsequence (LCS). It effectively measures the semantic overlap between the generated answer and the reference answer. The calculation formula is as follows:
\begin{equation}
{R_{lcs}}=\frac{LCS \left ({X,Y} \right)}{m}
\end{equation}
\begin{equation}
{P_{lcs}}=\frac{LCS\left ({x,Y}\right )}{n}
\end{equation}
\begin{equation}
{F_{lcs}}=\frac{\left ({1 + \beta^2}\right ){R_{lcs}}{P_{lcs}}}{{R_{lcs}}+ \beta^2{P_{lcs}}}
\end{equation}
where X denotes the reference answer (length m), Y denotes the predicted answer (length n), and LCS(X, Y) is the length of their longest common subsequence. The parameter $\beta$ is used to adjust the weight between precision and recall. In this study, we set $\beta$ = 1.2, slightly biased toward recall in order to meet the completeness requirement for answering legal questions.

Although traditional automatic evaluation metrics can quantify the surface similarity of texts, they are insufficient to comprehensively evaluate the professionalism, compliance, and practicality of legal answers. Therefore, this study introduces the “LLM as a Judge” evaluation framework, which uses a specialized large language model to perform semantic-level evaluation across multiple dimensions.

The evaluation dimensions include: Correctness: factual accuracy of the answer and correctness of legal references. Legality: compliance with current laws, regulations, and judicial practice. Completeness: full coverage of the core legal points related to the question. Clarity: clear expression, logical structure, and ease of understanding. Faithfulness: avoiding hallucinations and staying faithful to statutes and facts.

A weighted scoring method is adopted, with weights assigned according to importance in judicial contexts: 
\begin{equation}
FinalScore =\sum_{i= 1}^{5}{wi} \times {{score}_i}
\end{equation}
Correctness and Legality are assigned higher weights (0.25 each), while Completeness, Clarity, and Faithfulness are assigned moderate weights (0.15–0.20 each). The total score ranges from 0 to 1. In this study, Google Gemini-2.5-flash-lite is employed as the judge model to ensure professionalism and consistency in evaluation.

\subsection{Experimental Results}
\subsubsection{Performance Analysis of Multiple Baseline Models}
Experiment 1 aims to compare the performance of different baseline models (ChatGPT-4o, Qwen3-235B-A22B and DeepSeek-v3.1,) under three configurations: the baseline setup, the RAG model, and the hybrid model, on the Law\_QA validation set. This evaluation examines the gains introduced by RAG and multi-model ensembling. The result is shown in Table \ref{tab:1}.
\begin{table}[htbp]
    \centering
    \caption{Performance Comparison of Different Models under Baseline, RAG, and Hybrid Configurations}
    \resizebox{\columnwidth}{!}{
        \begin{tabular}{lccc}
            \hline
            Method                     & F1     & ROUGE-L  & LLM Judge \\
            \hline
            Baseline(GPT-4o)           & 0.2682 & 0.1875 & 0.883     \\
            RAG model(GPT-4o)          & 0.2740 & 0.1953 & 0.891     \\
            Hybrid model(GPT-4o)       & 0.2864 & 0.2103 & 0.914     \\
            \hline
            Baseline(Qwen3)            & 0.1923 & 0.1277 & 0.842     \\
            RAG model(Qwen3)           & 0.2235 & 0.1438 & 0.849     \\
            Hybrid model(Qwen3)        & 0.2434 & 0.1669 & 0.863     \\
            \hline
            Baseline(Deepseek-3.1)     & 0.3352 & 0.2341 & 0.934     \\
            RAG model(Deepseek-3.1)    & 0.3584 & 0.2501 & 0.953     \\
            Hybrid model(Deepseek-3.1) & 0.3612 & 0.2588 & 0.954     \\
            \hline
        \end{tabular}
    }
    \label{tab:model_performance}
\end{table}

Furthermore, across all models, the hybrid approach achieves the highest performance. For example, relative to the ChatGPT-4o baseline, the hybrid model improves F1 by +0.0182, ROUGE-L by +0.0228, and the LLM Judge score by +0.031. Similar gains are observed with the other baselines, highlighting that the combination of retrieval and multi-model ensembling consistently enhances robustness, completeness, and traceability. These findings demonstrate that our method generalizes well across different generative backbones, making it particularly suitable for judicial scenarios where precision of legal terminology and compliance with legal standards are critical.

\subsubsection{Ablation Study}
To further analyze the contributions of RAG and multi-model ensembling, Experiment 2 conducts ablation studies, comparing each module against the ChatGPT-4o baseline. Table 2 reports the results. The result is shown in Table \ref{tab:2}.


\begin{table}[htpb]
\centering
\caption{Ablation Study Results on RAG and Multi-Model Ensembling Contributions}
\label{tab:2}
\resizebox{\columnwidth}{!}{
  \begin{tabular}{c|cc|ccc}
  \hline
  \multicolumn{1}{c|}{Baseline} & \multicolumn{1}{c|}{RAG} & Multi-model & F1     & ROUGE-L & LLM Judge \\
  \hline
  \checkmark &   &   & 0.3352 & 0.2341 & 0.934 \\
  \hline
  \checkmark & \checkmark &   & 0.3584 & 0.2501 & 0.953 \\
  \hline
  \checkmark &   & \checkmark & 0.3440 & 0.2413 & 0.942 \\
  \hline
  \checkmark & \checkmark & \checkmark & 0.3612 & 0.2588 & 0.954 \\
  \hline
  \end{tabular}%
}
\end{table}

The ablation results show that the introduction of RAG alone improves F1 (+0.0232), ROUGE-L (+0.016), and LLM-as-a-Judge (+0.019), indicating that the retrieval of verifiable legal evidence improves semantic precision. The introduction of multi-model assembly alone yields only marginal improvements (F1 +0.0088, ROUGE-L +0.0072, Judge +0.008), suggesting that its effect is limited when retrieval coverage is lacking. When RAG and ensembling are combined, performance improves significantly, confirming their complementarity : RAG ensures traceability when retrieval succeeds, while ensembling compensates for knowledge-base gaps. Together, they enhance the overall reliability of legal QA.

\subsubsection{Analysis of Knowledge-Base Indexing Strategies}
Experiment 3 evaluates the impact of different knowledge-base indexing strategies on RAG performance, aiming to optimize retrieval hit rate and answer quality. Table 3 presents the comparison. The result is shown in Table \ref{tab:3}.


\begin{table}[htpb]
\centering
\caption{Performance Comparison of Different Knowledge-Base Indexing Strategies}
\label{tab:3}
\begin{tabularx}{\linewidth}{>{\raggedright\arraybackslash}m{2cm} *{3}{>{\centering\arraybackslash}m{1.5cm}}}
\hline
Indexing Strategy & F1 & ROUGE-L & LLM Judge \\
\hline
Question & 0.3217 & 0.2295 & 0.919 \\
\hline
Question + Answer & 0.3428 & 0.2426 & 0.923 \\
\hline
Question + Candidate Answer & 0.3584 & 0.2501 & 0.953 \\
\hline
\end{tabularx}
\end{table}

The three strategies include:(i) Question + Answer embedding (mixing question and answer for embedding). (ii) Question-only embedding. (iii)Question + Candidate Answer embedding, where the candidate answer is provided in the dataset. The candidate answer is semantically correct but expressed in a more conversational style than the standard answer, which improves retrieval hit rate when used as an additional embedding.
Results show that the Question + Candidate Answer strategy performs best (F1 = 0.3584, ROUGE-L = 0.2501, Judge = 0.953), surpassing both Question-only (+0.0367, +0.0206, +0.034) and Question + Answer embedding (+0.0156, +0.0075, +0.03). This demonstrates that incorporating candidate answers improves retrieval diversity and accuracy, enhancing RAG’s ability to handle varied user expressions in judicial scenarios.

\section{Conclusion and Future Work}
This paper proposes a hybrid legal QA agent tailored for judicial scenarios, which integrates retrieval-augmented generation (RAG) with multi-model ensembling to deliver legal consultation services characterized by high reliability, traceability, and dynamic updating. When user queries hit the knowledge base, the system applies RAG to ensure the accuracy of legal references; when no match is found, it generates candidate answers from multiple models, selects the optimal response through a scoring mechanism, and updates the knowledge base dynamically after human review. The experimental results indicate that our system achieves notable improvements over the baseline and standalone RAG methods, confirming its effectiveness in enhancing legal QA performance.

Looking forward, the system can be further extended to multi-modal forensics, privacy protection, and automated updating, and can be integrated with trusted execution environments and metaverse technologies to broaden the application of judicial AI in authenticity verification and security protection. At the same time, we call for greater research focus on AI safety and standardization in judicial contexts, so as to guarantee the reliability and fairness of such technologies in high-stakes domains.


\vspace {5mm}
\centerline{\bf{References}}
\begin{thebibliography}{99}
\bibitem[1]{re1} P. Lewis, E. Perez, A. Piktus, et al., “Retrieval-Augmented Generation for Knowledge-Intensive NLP Tasks,” in Advances in Neural Information Processing Systems (NeurIPS), 2020.
\bibitem[2]{re2} I. Chalkidis, T. Zhong, E. Fergadiotis, et al., “LexGLUE: A Benchmark Dataset for Legal Language Understanding in English,” ACL, 2022
\bibitem[3]{re3} Z. Ji, N. Lee, R. Frieske, et al., “A Survey of Hallucination in Natural Language Generation,” arXiv:2202.03629, 2024.
\bibitem[4]{re4} T. Wu, J. Lin, H. Chen, et al., “Continual Learning for Large Language Models: A Survey,” arXiv:2402.01364, 2024.
\bibitem[5]{re5} A. Asai, Z. Wu, Y. Wang, A. Sil, H. Hajishirzi, “Self-RAG: Learning to Retrieve, Generate, and Critique through Self-Reflection,” ICLR, 2024
\bibitem[6]{re6} J. Wang, B. Athiwaratkun, C.Z hang, J. Zou, “Mixture-of-Agents Enhances Large Language Model Capabilities,” arXiv:2406.04692, 2024.
\bibitem[7]{re7} L. Zheng, N. Guha, B. R. Anderson, P. Henderson, D. E. Ho, “When Does Pretraining Help? … the CaseHOLD Dataset,” ICAIL, 2021.
\bibitem[8]{re8} N. Guha, J. Nyarko, D. E. Ho, et al., “LegalBench: A Collaboratively Built Benchmark for Measuring Legal Reasoning in LLMs,” arXiv:2308.11462, 2023.
\bibitem[9]{re9} R. Goebel, Y. Kano, M. -Y. Kim, et al., “Overview of Benchmark Datasets and Methods for the Legal Information Extraction and Entailment Competition (COLIEE 2024),” 2024.
\bibitem[10]{re10} X. Yang, Y. Li, H. Qi, S. Lyu, “Exposing Deep Fakes Using Inconsistent Head Poses,” ICASSP, 2019.
\bibitem[11]{re11} H. Haliassos, R. Zhang, S. lbanie, P. Pérez, Y. Zisserman, “Lips Don’t Lie: A Generalisable and Robust Approach to Face Forgery Detection,” CVPR, 2021.
\bibitem[12]{re12} N. Wiratunga, S. Massie, A. Belkhouja, A. Ade-Ibijola, “Case-Based Reasoning Meets LLMs: A Retrieval-Augmented Legal QA Framework,” ICCBR, 2023.
\bibitem[13]{re13} X. Li, J. Zhao, T. Zhang, “LexRAG: A Benchmark for Retrieval-Augmented Legal Dialogue Systems,” LREC-COLING, 2024.
\bibitem[14]{re14} R. Barron, A. Santos, J. Costa, “Efficient Retrieval in Legal QA via Knowledge Graphs and NMF,” Artificial Intelligence and Law,
\bibitem[15]{re15} L. Zheng, K. Z. Chiang, D. Zeng, et al., “Judging LLM-as-a-Judge with MT-Bench and Chatbot Arena,” NeurIPS D\&B, 2023.
\bibitem[16]{re16} X. Wang, J. Wei, D. Schuurmans, et al., “Self-Consistency Improves Chain-of-Thought Reasoning in LMs,” arXiv:2203.11171, 2022.
\bibitem[17]{re17} X. V. Lin, X. Chen, M. Chen, et al., “RA-DIT: Retrieval-Augmented Dual Instruction Tuning,” arXiv:2310.01352, 2023.
\bibitem[18]{re18} G. Izacard, F. Petroni, L. Hosseini, et al., “ATLAS: Few-shot Learning with Retrieval-Augmented Language Models,” arXiv:2208.03299, 2022.
\bibitem[19]{re19} J. Johnson, M. Douze, H. Jégou, “Billion-Scale Similarity Search with GPUs,” arXiv:1702.08734, 2017.
\bibitem[20]{re20} J. Chen, S. Xiao, P. Zhang, K. Luo, D. Lian, and Z. Liu, “BGE M3-Embedding: Multi-Lingual, Multi-Functionality, Multi-Granularity Text Embeddings Through Self-Knowledge Distillation,” arXiv preprint arXiv:2402.03216, 2024
\bibitem[21]{re21} S. Gupta, R. Ranjan, S. N. Singh, “A Comprehensive Survey of Retrieval-Augmented Generation (RAG): Evolution, Current Landscape and Future Directions,” arXiv:2410.12837, 2024.
\bibitem[22]{re22} S. M. Qureshi, A. S. Malik, A. H. Abdullah, “Deepfake Forensics: A Survey of Digital Forensic Methods for Multimodal Deepfake Identification on Social Media,” Multimedia Tools and Applications, 2024.
\bibitem[23]{re23} J. Qi, G. Sarti, R. Fernández, A. Bisazza, “Model Internals-based Answer Attribution for Trustworthy RAG,” arXiv:2406.13663, 2024.
\end{thebibliography}
\end{document}